\documentclass[sigconf]{aamas} 

\newcounter{example}[section]
\newenvironment{example}[1][]{\refstepcounter{example}\par\medskip
   \noindent \textbf{Example~\theexample. #1} \rmfamily}{\medskip}

\usepackage{balance} 

\usepackage{algorithm}
\usepackage{algorithmic}

\usepackage{amsmath}
\usepackage{multirow}
\usepackage{multicol}

\setcopyright{none}
\acmConference[ALA '23]{Proc.\@ of the Adaptive and Learning Agents Workshop (ALA 2023)}
{May 29-30, 2023}{London, UK, \url{https://alaworkshop2023.github.io/}}{Cruz, Hayes, Wang, Yates (eds.)}
\copyrightyear{2023}
\acmYear{2023}
\acmDOI{}
\acmPrice{}
\acmISBN{}
\acmSubmissionID{35}

\title[AAMAS-2023]{Strategy Extraction in Single-agent Games}

\author{Archana Vadakattu}
\affiliation{
  \institution{The University of Melbourne}
  \city{Melbourne}
  \country{Australia}}
\email{vadakattua@unimelb.edu.au}

\author{Michelle Blom}
\affiliation{
  \institution{The University of Melbourne}
  \city{Melbourne}
  \country{Australia}}
\email{michelle.blom@unimelb.edu.au}

\author{Adrian R. Pearce}
\affiliation{
  \institution{The University of Melbourne}
  \city{Melbourne}
  \country{Australia}}
\email{adrianrp@unimelb.edu.au}

\begin{abstract}

The ability to continuously learn and adapt to new situations is one where humans are far superior compared to AI agents. We propose an approach to knowledge transfer using behavioural strategies as a form of transferable knowledge influenced by the human cognitive ability to develop strategies. A strategy is defined as a partial sequence of events -- where an \textit{event} is both the result of an agent's action and changes in state -- to reach some predefined event of interest. This information acts as guidance or a partial solution that an agent can generalise and use to make predictions about how to handle unknown observed phenomena. As a first step toward this goal, we develop a method for extracting strategies from an agent's existing knowledge that can be applied in multiple contexts. Our method combines observed event frequency information with local sequence alignment techniques to find patterns of significance that form a strategy. We show that our method can identify plausible strategies in three environments: Pacman, Bank Heist and a dungeon-crawling video game. Our evaluation serves as a promising first step toward extracting knowledge for generalisation and, ultimately, transfer learning.
\end{abstract}


\newcommand{\BibTeX}{\rm B\kern-.05em{\sc i\kern-.025em b}\kern-.08em\TeX}

\begin{document}


\pagestyle{fancy}
\fancyhead{}


\maketitle 


\section{Introduction}
Humans have the natural ability to develop strategies in one context that can be generalised and applied in other contexts, providing a useful starting point to form a complete course of action in unfamiliar settings \cite{dapge:18,cf:13}. Existing work on transfer learning involves learning low-level information from a source task \cite{twsm:05,kb:07,bhtn:15}, which can be transferred to one or many target tasks. However, these approaches face limitations such as being restricted to certain classes of tasks due to learnt knowledge being insufficiently generalisable \cite{sm:22}. For instance, an agent may learn a possible solution to move a block in a video game. However, it cannot autonomously recognise the similarities to apply the solution in the real world. The agent would require external guidance to align the action or state spaces between the two environments. Suppose artificial agents can perform strategy synthesis, allowing them to utilise their existing knowledge better to continuously learn and react to novel situations. This would significantly improve an agent’s learning capabilities to handle a wide range of tasks with limited data where it can apply generalised strategies to guide behaviour.  

In this work, we propose a method of knowledge transfer in AI agents based on the human cognitive ability to develop strategies. As a first step, we seek to obtain transferable knowledge, in the form of behavioural strategies, from an agent’s past behaviour in other environments. The idea is that strategies could be used by the agent in contexts when the action spaces, reward functions or environment dynamics are different from those where the strategies originated, where recognising similarities requires a deeper, more abstract-level understanding of situations. 

A strategy is an abstract task structure representing a partial plan to achieve a goal in some environment. We adopt an intuitive definition of a strategy rather than the game theoretical definition, as referred to in multi-agent studies, which implies a complete set of instructions specifying how a player should make decisions \cite{pw:02}. A partial plan refers to the usability of the solution -- it may not completely solve a problem, but it provides a base for a complete solution. Our definition encompasses plans with a partial ordering of events as well as plans with potentially unnecessary events. For example, in the game of Pacman, a strategy may be to ``collect a power-up to defeat a ghost''. This could be abstracted to ``collect an item to defeat an enemy''. Defeating enemies is a common objective that appears in various games, making this strategy applicable in many contexts.

In this paper, we form strategies by extracting information from trajectories gathered in single-agent video game environments. Games inherently have strategies in order to achieve a range of goals that vary in complexity. More specifically, we consider a family of problems that use symbolic event representation to generate interpretable results. Trajectories are represented by a series of events, where an \textit{event} is defined as both the result of an agent's action and changes in state. Existing benchmark environments, such as those from the Arcade Learning Environment \cite{bnvb:13}, are reimplemented to generate events since the default game trajectories consist only of image-based state representations and low-level actions. 

The key contribution is our unique approach to strategy extraction by treating the problem as a sequential pattern mining task. The novelty lies in using a sequence alignment technique known as the Smith-Waterman algorithm \cite{sw:81} which is more commonly known for comparing DNA string sequences. We apply a modified Smith-Waterman algorithm combined with observed event frequencies across a dataset of trajectories to find patterns of significance that form a strategy. We consider case studies based on single-player video game environments to demonstrate the approach and develop performance benchmarks. The proposed method has broader applicability in single-agent environments, and games have been chosen for demonstration purposes. Our evaluation serves as a promising first step towards efficient and robust generalisation and how to best utilise the generalised strategies in new tasks and complex domains. Strategy generalisation and transfer are left as future work.

\section{Related Work}

Defining a reliable method for generalisation and transfer across multiple tasks is an open research problem. Multiple works have proposed transferring learned skills to solve abstract subtasks that are repeated in various games. \citet{ah:07} propose a method that transfers policies for achieving common sub-goals between tasks used to construct a model for solving the new task. Their method uses the options framework, a notable hierarchical reinforcement learning framework \citep{sps:99}. Options are learned from extracted sub-goals and stored in the agents' memory in the form of a policy, termination condition and an input set. The restriction of using knowledge transfer via options without abstraction means that transfer learning only works between agents which use the same learning algorithms. Learning abstract options is also challenging and will depend on the state and action space of the target task. In contrast, the form in which we represent our strategies can be flexible, depending on the type of learning agent. 

Strategy extraction is more commonly seen as a data-mining problem, typically demonstrated in real-time strategy (RTS) games which provide a testbed for many different strategy-based tasks \cite{shbs:04, btbrk:13, cecbtc:15, elnr:15}. There are many studies in sequence modelling and pattern mining that leverage information gathered from the analysis of state-action trajectories. Such techniques help to discover patterns in trajectories related to specific tasks. This information can inform our strategy construction. 

Most existing work focuses on finding strategies with the purpose of player modelling for applications such as learning an opponent's strategy, understanding a player's behaviour and finding a winning strategy \cite{elnr:15, sbkk:20}. The types of strategies typically modelled are complete solutions such as telling the player exactly how to reach the goal. Several extraction procedures extract strategies based on pattern frequency or \textit{support}; the percentage of sequences in a dataset that contain the pattern. For example, \citet{cecbtc:15} identify closed frequent patterns whose frequency is above some pre-specified support threshold, \citet{lrkp:13} detect significant sequential patterns based on the frequency of pattern prefixes and \citet{btbrk:13} use a combination of pattern prefix and sequential pattern frequency. The types of strategies obtained, although useful, do not provide any information at the event level in relation to each events' significance in achieving the goal. Subsequently, even if the game- or implementation-specific content is abstracted, these complete strategies, or plans, are highly unlikely to be applicable in other games. In the Pacman example, this would be like including individual movements and collection of dots in the strategy to kill a ghost. This in turn makes it difficult to generalise the extracted strategy as it will include many irrelevant events. 

\citet{shbs:04} use a text-based method, latent semantic analysis, to detect semantic information which they interpret as a game strategy. Inspired by this idea, our proposed approach uses the sequential pattern mining method known as \textit{local sequence alignment} to find useful causal information which we can use to form strategies. Local sequence alignment is a variant of global sequence alignment. The key difference is that the global variant treats sequences as a whole and does not identify partial sequential matches. A notable method for local sequence alignment is the Smith-Waterman algorithm, first proposed in 1981 by F. Smith and M. S. Waterman. We use this method, more commonly known for its application in bioinformatics to compare strings of nucleic acid or protein sequences, to compare game trajectories.

\section{Preliminaries}
 
We consider an agent that has been trained to play a game through reinforcement learning. A single-player game environment consists of a set of states $S$, available actions $A$ and rewards $R$. At each timestep, $t$, the player performs an action $a_t$ from the set $A$ and receives some reward $r_t$. A game trajectory, $\tau$ is a finite sequence of consecutive actions, events, rewards and observed states of the environment before and after the action is taken, $s_{t-1}$ and $s_t$ respectively. A \textit{subtrajectory} is a trajectory containing a subset of the elements from another trajectory, maintaining the same temporal ordering. A subtrajectory may not be equivalent to a subsequence; elements from the original trajectory can be dropped. For example, given a trajectory of the form ${a,b,c}$, a valid subtrajectory is ${a,c}$ as well as ${a,b}$ and ${b,c}$.

\subsection{Environments}
We demonstrate our method on three custom environments, implemented in OpenAI Gym \cite{bcpsstz:16}. All three environments have discrete action spaces. These environments are implemented as text-based versions where all aspects of the environments are represented by ASCII characters. We use Proximal Policy Optimisation (PPO) \cite{swdrk:17} to learn a policy for each environment. Image-based environments could also be used as long as the agent's behaviour and environment dynamics can be captured in a sequence of symbolic events. 

\subsection{Pacman} 
\begin{figure}[t]
\centering
\includegraphics[width=0.9\columnwidth]{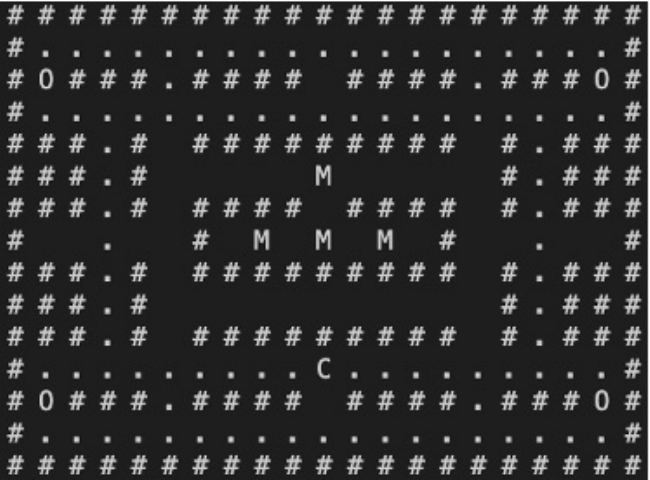} 
\caption{Example Pacman game state: $\mathrm{C}$ denotes Pacman's location; $\mathrm{M}$ a ghost; $\#$ a wall; $\mathrm{O}$ a power-up; and $.$ a dot.}
\label{fig:1}
\end{figure}

We developed a version of Pacman inspired by the original Atari game, as a 2D environment, as shown in Figure \ref{fig:1}. The agent's action space contains move actions in four directions: up, down, left and right. The following events can take place in the environment: \textit{move, collect power-up/dot} and \textit{kill a ghost}. An observation of the game state is a grid where each cell contains a numeric value representing a game component -- walls, ghosts, dots, power-ups and Pacman. Pacman kills a ghost by collecting a power-up and then moving over the ghost within the next five movements.

During training, rewards are received when the agent collects items (dots and power-ups). Additionally, the agent is rewarded for killing ghosts, with doubled rewards received when killing ghosts in succession. The map topology, and the starting locations of the ghosts, remain unchanged for every episode of training. Ghosts move around the map during the game. 

\subsection{Dungeon Crawler} Dungeon Crawler is an exploration game where the goal of the agent is to navigate through a maze to collect a key and then escape through a door. The agent must avoid monsters in its search for the key or kill the monsters by collecting a weapon. Points are rewarded for collecting items (weapons and keys), killing monsters and escaping through the door. Similar to the Pacman environment, the available actions are up, down, left and right. The following events can take place in the environment: \textit{move, collect weapon (gun), collect weapon (sword), collect key, kill a monster} and \textit{unlock door}. 

\begin{figure}[t]
\centering
\includegraphics[width=\columnwidth]{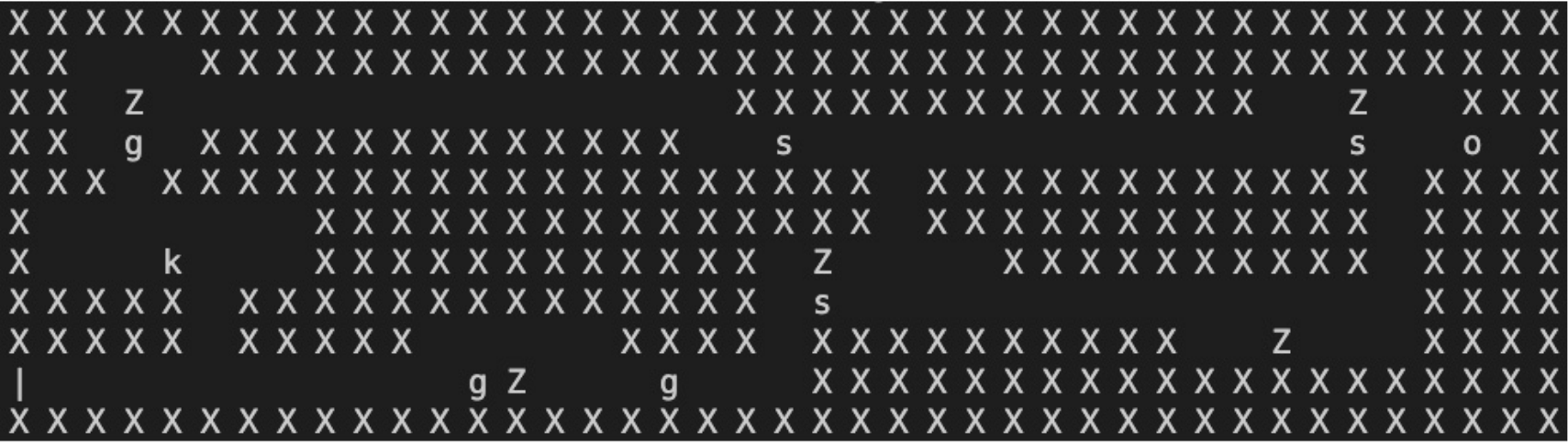} 
\caption{Example Dungeon Crawler game state: $\mathrm{o}$ denotes the location of the player; $\mathrm{Z}$ a monster; $X$ a wall; ${|}$ the door; $\mathrm{k}$ the key; and $g$ and $s$ the gun and sword respectively.}
\label{fig:2}
\end{figure}

The dungeon map is fixed for every training episode, however the items and monsters are randomly placed (6 monsters and 3 of each weapon type). An example of an initial game grid is shown in Figure \ref{fig:2}. An observation of the game state contains the shortest-path distances between the agent's current location to various entities (weapons, monsters, the key and the door). 

There are multiple events for which we expect to find strategies. The first is the key-door collection strategy. The second is the weapon-monster killing strategy. Multiple weapon types are available (e.g. `gun' and `sword'). Two valid strategies in this context may be ``collect the gun to kill a monster", and ``collect the sword to kill a monster". 

\subsection{Bank Heist} 
\begin{figure}[t]
\centering
\includegraphics[width=0.9\columnwidth]{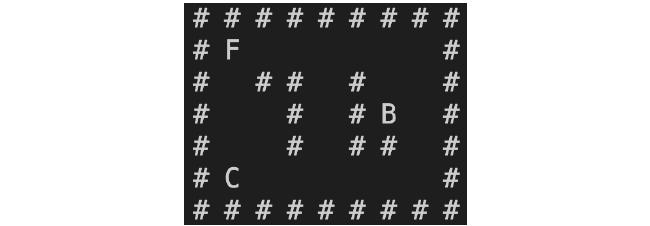} 
\caption{Example Bank Heist game state: $\mathrm{C}$ denotes the location of the player's car; $\mathrm{B}$ a bank; $\#$ a wall; and $\mathrm{F}$ a fuel tank.}
\label{fig:3}
\end{figure}

Influenced by the Atari version of the same name, we developed a simple Bank Heist game as shown in Figure \ref{fig:3}. In this version, the agent must rob five banks without getting caught by police cars or running out of fuel. The agent begins with a full fuel tank which then depletes through movement and dropping dynamite. Collecting a fuel tank refills the tank to 100\%. The agent may drop dynamite to destroy police cars however dropped dynamite can also kill the agent. Fixed-value rewards are given when the agent destroys a police car and collects a fuel tank. The reward earned for robbing a bank increases for each bank that is robbed. 

The environment starts with one bank and a fuel tank. Robbing banks causes more banks to appear in random empty spaces. An observation of the game state is a list of values for each available action containing a numerical representation of how beneficial the action is expected to be if taken in the next step. 

\subsection{Local Sequence Alignment}

Suppose we have sequences $A=a_0, a_1, ..., a_m$ and $B=b_0, b_1, ..., b_n$ where $m$ and $n$ are the respective sequence lengths. Local sequence alignment can be applied to compare $A$ and $B$ and find regions where the two sequences are similar to each other. The result is one or more subsequences deemed to be most similar according to some metric.

The Smith-Waterman algorithm performs local sequence alignment by first forming a scoring matrix, populated by comparing the $m$ elements of $A$ with the $n$ elements of $B$. The matrix dimensions are $(m+1) \times (n+1)$ and the first row and column are filled with zeros. The matrix cell $(i,j)$ where $i \in [1, m+1]$ and $j \in [1, n+1]$ is assigned a score based on the comparison of $A_i$ and $B_j$, and the values in adjacent matrix cells. In the original algorithm, these scores are computed as per Equation \ref{eq:1}, where $s$, $d$ and $g$ are user-defined \textit{match}, \textit{mismatch} and \textit{gap} scores respectively. 
\begin{equation}\label{eq:1}
    H_{AB}(i,j) = max
    \begin{cases}
        0 \\
        H_{AB}(i-1,j-1) + s, \,\text{if } A_i = B_j \\
        H_{AB}(i-1,j-1) + d, \,\text{if } A_i \neq B_j \\
        H_{AB}(i,j-1) - g \\
        H_{AB}(i-1,j) - g 
    \end{cases}
\end{equation}

When filling out the values in the aforementioned matrix, Smith-Waterman keeps track of which adjacent cell was used to compute the score for cell $(i,j)$. This score, $H_{AB}(i,j)$, will use at most one of $H_{AB}(i-1,j-1)$, $H_{AB}(i,j-1)$, and $H_{AB}(i-1,j)$. This knowledge is later used, through a \textit{traceback} procedure, to form the output of the sequence comparison -- a similar subsequence. 

The traceback process starts from the highest-scoring cell, $(i,j)$, including either $A_i$ or $B_j$ in our output subsequence. The method then recalls which adjacent cell was used to calculate the value in $(i,j)$, and moves to that cell. For each diagonal movement to a cell $(k,l)$, one of $A_k$ and $B_l$ is placed at the front of the evolving output subsequence. For vertical or horizontal movements, a \textit{gap} token is added. This process continues until a cell with a zero is reached.

We adapt Smith-Waterman to perform pairwise comparisons on trajectories and return a subtrajectory. Given our motivation for comparing trajectories is to highlight important elements, we have made several changes to the original scoring function shown in Equation \ref{eq:1}. 

First, we implement a custom function for element comparison. The implementation depends on the choice of environment and which characteristics are available. For the environments previously described which output event descriptions in the form of strings, this function performs string comparisons. Second, we maintain match and mismatch scoring since we want to maintain elements that exist in both trajectories and de-emphasize all others. Third, we discard penalties ($g$) for the presence of gaps in the resulting subtrajectory. Smith-Waterman aims to find a similar \textit{subsequence} when comparing two sequences, explicitly placing gap tokens between consecutive items in the result if they do not occur consecutively in the two sequences being compared. We are interested in identifying the important similarities (events) between two trajectories, and their relative temporal ordering. We do not care whether gaps should be placed in the resulting subtrajectory, or how many, in the context of our strategy extraction objective. 

Finally, we include a weighting, $\mathcal{W}$, capturing additional information about the elements being compared, beyond whether they match, to further influence which elements appear in our output subtrajectory. We describe in subsequent sections how we instantiate $\mathcal{W}$ when comparing trajectories. These changes are reflected in Equation \ref{eq:2}. 
\begin{equation}\label{eq:2}
    H_{AB}(i,j) = 
    \begin{cases}
        H_{AB}(i-1,j-1) + s \ \mathcal{W}, \,\text{if } A_i = B_j \\
        H_{AB}(i-1,j-1) + d \ \mathcal{W}, \,\text{if } A_i \neq B_j
    \end{cases}
\end{equation}

Once the scoring matrix is filled, we traceback through the matrix to discover which of its cells will be used to determine the output, following the procedure described earlier. When deciding on whether to include $A_i$ or $B_j$ in our resulting subtrajectory, after a diagonal movement to cell $(i,j)$, we choose the event from the shorter trajectory.

\begin{figure}[t]
\centering
\includegraphics[width=\columnwidth]{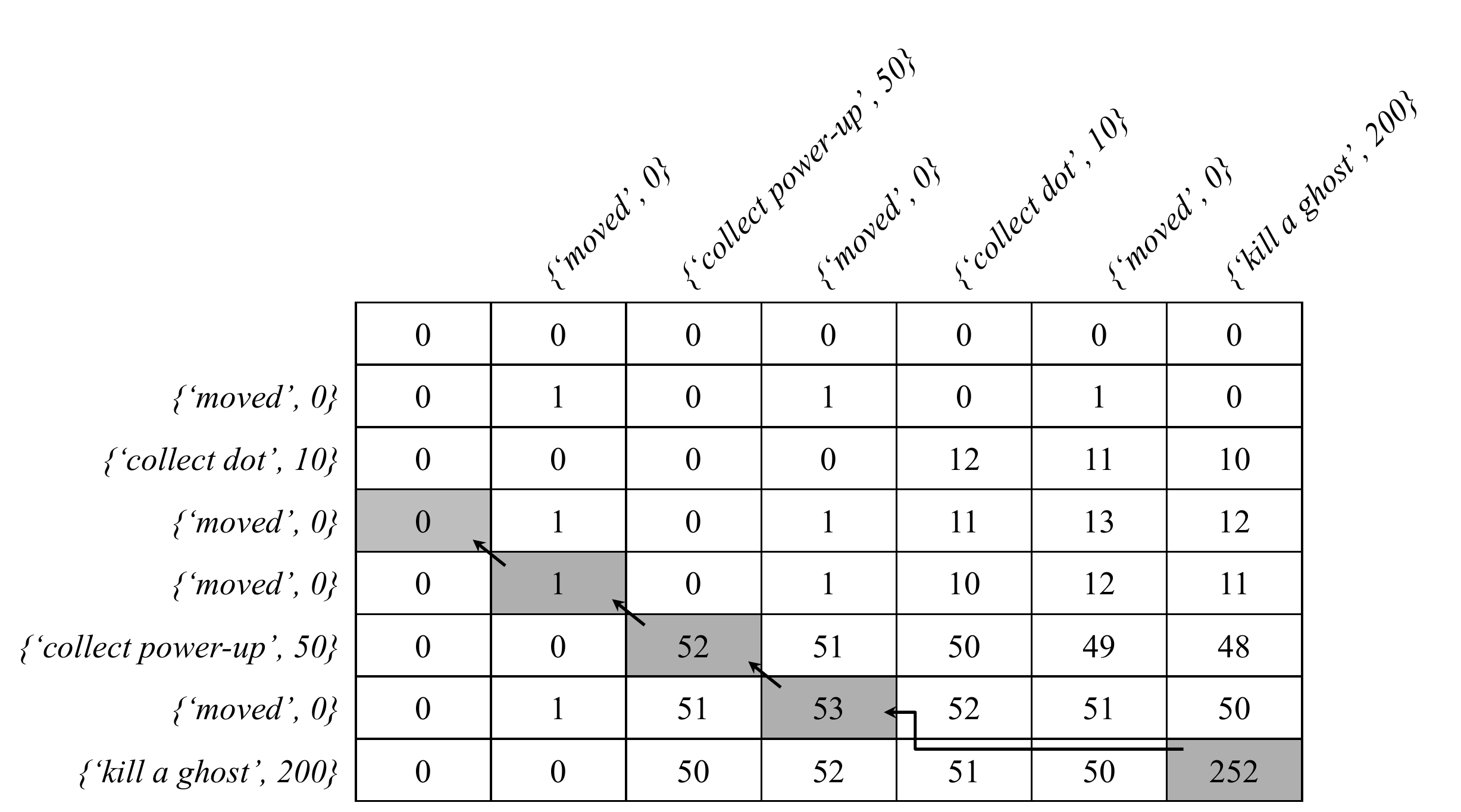} 
\caption{Example Smith-Waterman scoring matrix constructed using Equation \ref{eq:2} where $s=1$, $d = -1$ and $\mathcal{W}=max(1, reward(A_i), reward(B_j))$. Trajectory elements from $A$ and $B$ are displayed in terms of their events and corresponding rewards. The directional arrows depict the recursive traceback process, highlighting the cells which are used to determine the output subtrajectory (shaded).}
\label{fig:4}
\end{figure}

Figure \ref{fig:4} shows an example of a scoring matrix that has been constructed for two Pacman trajectories, excluding state and reward information for simplicity, $A=$\{``move", ``collect dot", ``move", ``move", ``collect power-up", ``move", ``kill a ghost"\} and $B=$\{``move", ``collect power-up", ``move", ``collect dot", ``move", ``kill a ghost"\}. In this example, $\mathcal{W}$ is instantiated using received rewards for the purpose of demonstration. Formally this can be written as $\mathcal{W}=max(1, reward(A_i), reward(B_j))$, where $reward(A_k)$ denotes the reward received by the agent after the event $A_k$ in the relevant trajectory. 
The output subtrajectory for this example is \{``move", ``collect power-up", ``move", ``kill a ghost"\}, including events from the shorter of the two trajectories, $B$. Since the rewards may not always reflect the importance of an event in a strategy, we require a more robust definition of $\mathcal{W}$. Later, we will define $\mathcal{W}$ based on observed event frequencies (or likelihoods).

We use the Smith-Waterman algorithm for three reasons. First, as a local, as opposed to global, sequence alignment method, it allows subsequence similarities to be detected. Second, it supports the addition of gaps in the output subsequences to replace one or more sequence elements. While we do not penalise the presence of gaps, or explicitly add gap tokens to our subtrajectories, gap allowance is important. In a trajectory, the events that are important to a goal may not always occur consecutively. For example, in Pacman, we want to find strategies like ``collect a power-up, kill a ghost" rather than explicitly list all the moves and dot collections that occur in between. Finally, the function and metrics used for element comparisons and score assignment are flexible allowing for customisation.

\section{Method}

\begin{figure*}[t]
\centering
\includegraphics[width=0.7\textwidth]{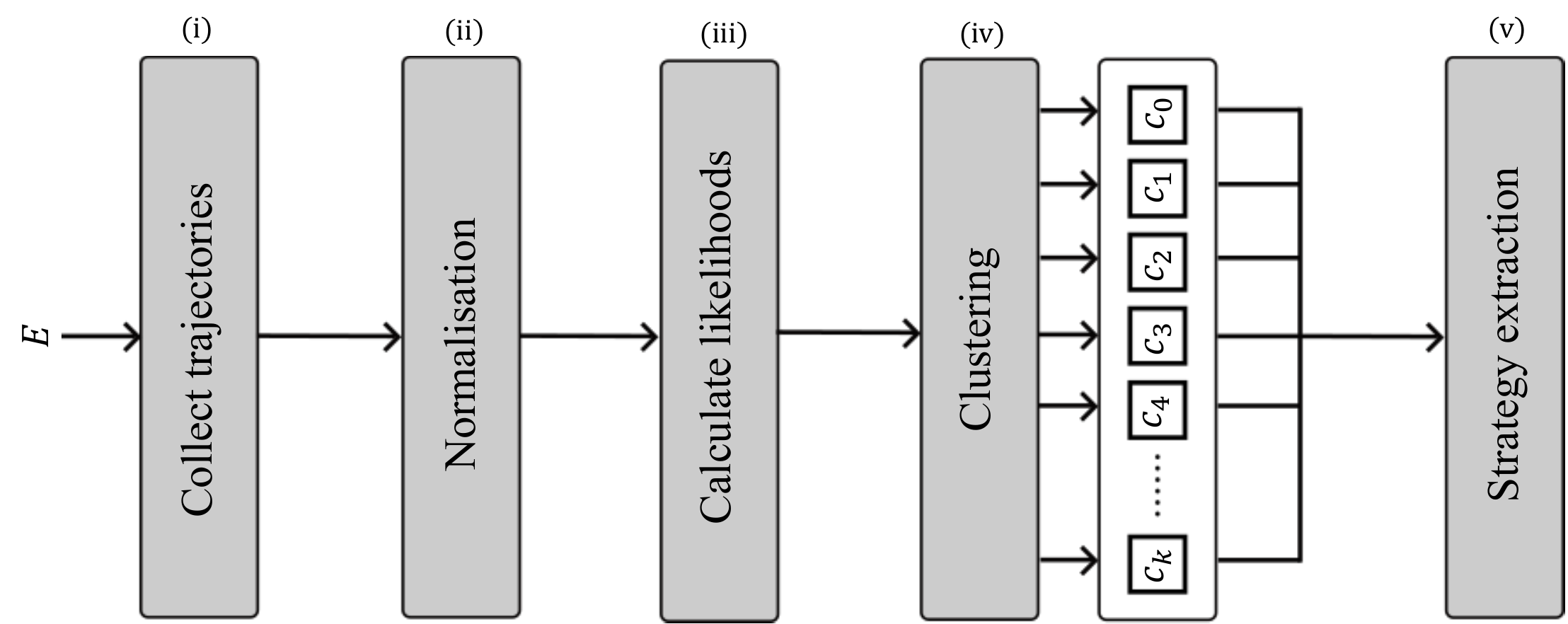} 
\caption{The stages of our strategy extraction approach for an event of interest $E$. Details are as follows: $(i)$ generate positive and negative trajectories for $E$ through simulation; $(ii)$ using the positive and negative trajectory length distributions, normalise the negative trajectories; $(iii)$ determine likelihood values for events being part of a strategy; $(iv)$ create clusters from the positive trajectories; and $(v)$ for each cluster, perform strategy extraction to obtain candidate strategies through pairwise comparison.}
\label{fig:5}
\end{figure*}

In this section, we describe each stage of our proposed strategy extraction approach (Figure \ref{fig:5}). Given a policy $\pi$ for playing a game, $\mathcal{G}$, our approach first identifies \textit{events of interest}. These events occur when playing $\mathcal{G}$, and represent goals or sub-goals that an agent may wish to achieve.

For an event of interest, $E$, our approach collects two sets of trajectories. The first set contains trajectories in which $E$ occurs, denoted positive trajectories, and the second trajectories in which $E$ does not occur, denoted negative trajectories. The lengths of the negative trajectories are normalised so that the distribution of trajectory lengths of the positive and negative sets match. The positive and normalised negative trajectories are then used to compute the likelihood of each possible event being part of a strategy for achieving $E$. The likelihood of an event is computed by comparing the frequency with which it appears in the positive trajectories relative to the negative. Events that appear more often in positive trajectories are more likely to be part of a strategy.

Pairs of positive trajectories will ultimately be compared using our Smith-Waterman adaptation. To reduce the complexity of these comparisons, we remove events whose likelihood falls below a threshold from all positive trajectories. To select pairs of trajectories for comparison, we first cluster trajectories on the basis of the events they contain. We sample pairs of trajectories from each cluster, performing pairwise comparisons using the adapted Smith-Waterman algorithm. The result for each cluster is a set of candidate strategies for achieving $E$. The purpose of clustering trajectories on the basis of their events is to ensure that we identify different strategies for achieving the same $E$, when present.

\subsection{Discovering Events of Interest}
We first discover events that occur in the environment that the player could use a strategy to achieve. We use a method of detection based on rewards, however this can be replaced with more sophisticated approaches that do not rely solely on the reward function \cite{ah:07,pvr:19}. 

Using the policy $\pi$, we simulate the PPO agent in the environment for $N$ episodes and gather trajectories from which events of interest are identified. The first $N/2$ episode trajectories are used to obtain an estimate of the average episode reward $r_{avg}$. The remaining $N/2$ episodes are simulated, searching for events that receive a reward greater than $r_{avg}$. These events form our \textit{events of interest}.

\begin{example}
    \textbf{(Pacman)} \textit{We collect $N = 100$ trajectories to discover events of interest. Across the first half, $r_{avg} = 11$. Across the second half, events that receive a reward greater than $11$ include ``collect power-up" (reward = 50) and ``kill a ghost" (reward = 200, 400).}
\end{example}

\subsection{Collecting Trajectory Data}
For a selected $E$, trajectories are collected to form a dataset for strategy extraction. When the PPO agent is simulated in the environment, historical trajectories $\tau_H$ are saved. Positive trajectories, $\tau_P$, are collected when the agent reaches an event of interest mid-simulation. The agent's trajectories from the beginning of the episode to the event of interest are saved. To collect negative trajectories, $\tau_N$, we simulate a random agent in the environment, one which selects a random (valid) action at each step. From the trajectories obtained from these simulations, a random set of entire-episode trajectories is selected, filtering out the trajectories which contain the event of interest. 

Our approach relies on collecting the same number of positive and negative samples. Suppose this becomes too difficult; for example, if the policy $\pi$ is trained sufficiently well with respect to a given event of interest such that all trajectories in $\tau_H$ are positive. Using a random agent, or any other basic heuristic, to guide the agent behaviour when generating negative trajectories, accommodates this case.

\begin{example}
    \textbf{(Pacman)}  \textit{When $E=$``kill a ghost", one possible trajectory from $\tau_P$ is} \{\textit{``move", ``collect dot", ``move", ``collect power-up", ``move", ``kill a ghost"}\}. \textit{A trajectory from $\tau_N$ is} \{\textit{``move", ``collect dot", ``move", ``collect dot", ``move"}\},  \textit{at which point Pacman is killed by a ghost}. 
\end{example}

\subsection{Normalisation}

Depending on the environment and event of interest, there could be significant variation between the length distributions of $\tau_P$ and $\tau_N$. For the chosen $E$, negative trajectories may be longer on average than positive trajectories. In such cases, this obscures the useful information that we wish to gather from comparing the two sets in later stages. Significantly longer negative trajectories influence our calculation of event likelihoods, as these are based on the relative frequency with which specific events occur across the two sets. 

To solve this problem, $\tau_N$ is normalised to ensure its length distribution matches that of $\tau_P$. 
We randomly sample a length from the length distribution of $\tau_P$, $l_{\tau_P}$, and also sample a longer trajectory from $\tau_N$, $t_N$. A subsequence of the desired length is extracted from $t_N$ and added to the new normalised set, $\tau_N\prime$. We repeat this process until $\tau_N\prime$ contains the same number of trajectories as $\tau_P$.

\begin{figure}[t]
\centering
\includegraphics[width=0.9\columnwidth]{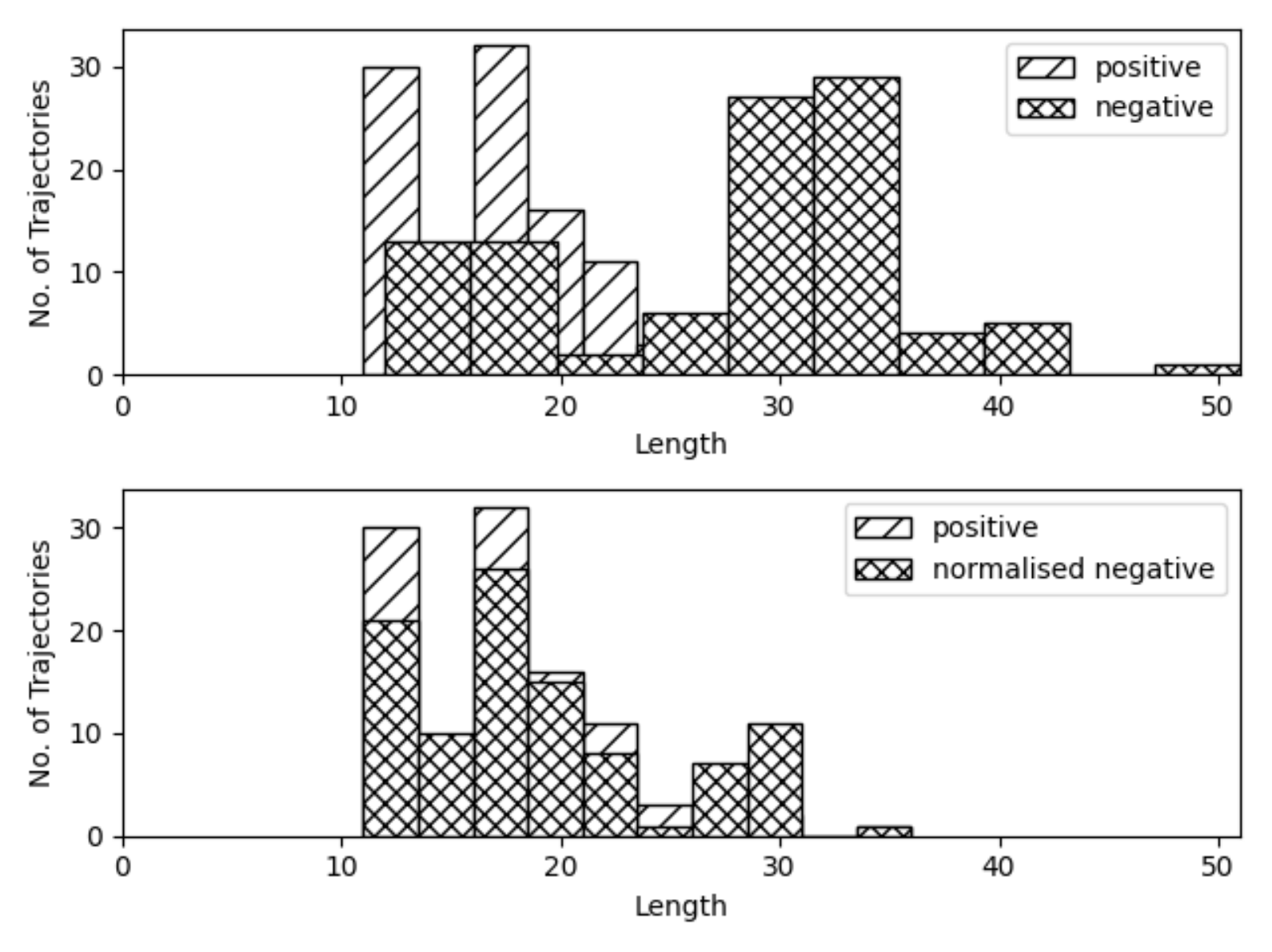} 
\caption{Before and after normalising the length distributions of the negative trajectories obtained from simulating the agent in a small (10 x 6) Pacman environment. Positive trajectories include the event ``kill a ghost". }
\label{fig:6}
\end{figure}

This normalisation step is not applied if the length distributions are equal, in terms of their respective mean and standard deviation, or if the negative trajectories are, on average, shorter than the positive trajectories. Figure \ref{fig:6} shows an example of applying our normalisation method on trajectories obtained from agent simulations in the Pacman environment. 

\begin{example}
    \textbf{(Pacman)} \label{sssec:ex3} \textit{We expect a strategy for killing a ghost to include ``collect power-up". This event is in $100\%$ of trajectories in $\tau_P$. We observe that episodes in which the agent does not kill a ghost involve long event sequences reflecting random exploration, item collection or ghost avoidance. As a result, we observe that ``collect power-up" also appears in $100\%$ of the negative trajectories. After normalisation, fewer of the negative trajectories contain this event.}
\end{example}

\begin{algorithm}[!tb]
\caption{Find Strategies for an Event of Interest}
\label{alg:1}
\textbf{Input}: $\mathcal{C}$ (clusters of trajectories), $\mathcal{L}$ (event likelihoods)\\
\textbf{Output}: Strategy set, $\mathcal{S}$
\begin{algorithmic}[1] 
\STATE Let $\mathcal{S} = \emptyset$.
\FOR{$c \in \mathcal{C}$}
    \STATE{$t_c = ShortestTrajectory(c)$}
    \FOR{$t_j \in c \setminus\{t_c\}$}
        \STATE{$result \gets SmithWaterman(t_c, t_j, \mathcal{L})$}
        \STATE{$\mathcal{S} \gets \mathcal{S} \cup \{result\}$}
    \ENDFOR
\ENDFOR
\STATE \textbf{return} $\mathcal{S}$
\end{algorithmic}
\end{algorithm}

\subsection{Calculating Event Likelihood}
We have assumed that there is an equal likelihood for every event to appear in one or more strategies however realistically, this is very unlikely. By analysing groups of trajectories, we gain insights into the distribution of events and recognise those which are more likely to be in a strategy. We derive likelihood values for each possible event and use this information to disregard events from the trajectories; simplifying them for subsequent calculations. 

To calculate the likelihood of an event, $e$, we compare its frequency of occurrence within $\tau_P$ and $\tau_N\prime$. If $e$ occurs considerably more often in the positive samples than in the negative samples, its likelihood of appearing in the strategy is very high. Conversely, if it appears more in the negative than the positive, then commonsense reasoning tells us that $a$ is unlikely to be part of the strategy. This is also true if the event appears approximately the same amount in both sets.

Likelihood is defined as the difference between occurrence frequencies of an event in each of $\tau_P$ and $\tau_N\prime$, denoted $f_{\tau_P}(e)$ and $f_{\tau_N\prime}(e)$ respectively. By default, all events have a likelihood of 1. If an event only appears in the negative trajectories and not in the positive trajectories, its likelihood is set to 0. Equation \ref{eq:3} formalises the likelihood calculation for an event $e$.
\begin{equation}
\label{eq:3}
    l_e = 
    \begin{cases}
        max(0, f_{\tau_P}(e)-f_{\tau_N\prime}(e)) & \text{if $e$ in $\tau_P$ and $\tau_N$} \\  
        1 & \text{if $e$ not in $\tau_N$} \\
        0 & \text{if $e$ not in $\tau_P$} \\
    \end{cases}
\end{equation}

Low-likelihood events are then removed from $\tau_P$ using a predefined threshold. We denote these modified trajectories as $\tau_P\prime$ which is used for the remaining stages. A threshold of 0.1 is used to obtain $\tau_P\prime$ across all experiments in this paper. 

\begin{example}
    \textbf{(Pacman)} \textit{If the event ``collect power-up" appears in $100\%$ of positive trajectories and $20\%$ in the negative trajectories; then its likelihood $l_{power-up} = (100 - 20) \div 100  = 0.8$. Computing likelihoods for the remaining events, we obtain the following values: ``move": 0, ``collect dot": 0.1, ``collect power-up":0.8, ``kill a ghost": 1.}
\end{example}

\subsection{Clustering}

In this stage, we create clusters from the trajectories in $\tau_P\prime$ in order to group trajectories that are related. A new cluster is created for each event that occurs in a trajectory in $\tau_P\prime$, and all trajectories that contain this event are added. In this way, a trajectory may appear in more than one cluster. We denote the set of generated clusters, $\mathcal{C}$.

The aim of clustering is to ensure we uncover patterns when there are many different versions of trajectories that achieve the same $E$. For example, in the Dungeon Crawler environment for $E=$``kill a monster", $\tau_P\prime$ may include trajectories that contain one event $e_1$ and not another $e_2$ and vice versa. We can create two clusters; one for trajectories that contain $e_1$ and another for $e_2$. In the final stage, each cluster is considered separately to find all possible strategies.

\subsection{Strategy Extraction}
In the final stage, the Smith-Waterman algorithm is used to perform multiple pairwise comparisons between trajectories in each cluster. 
Recall from Equation \ref{eq:2}, the weighting $\mathcal{W}$. When comparing trajectory elements $A_i$ and $B_j$, we use the weighting $\mathcal{W}=max(l_{A_i}, l_{B_j})$. Our choice to use likelihood in this way, rather than relying on rewards, is due to the large variance in the way rewards are assigned in different environments. If our method used rewards, the capability to extract appropriate strategies becomes dependent on the reward scheme of the game. We use likelihood values for strategy extraction to ensure the method is game-agnostic.

Algorithm \ref{alg:1} outlines our method for finding strategies for a given $E$.
For each cluster, the shortest trajectory (i.e., with the least number of events) is selected. A pairwise comparison between this trajectory, and all others in the cluster, is performed using the Smith-Waterman algorithm. The output of each pairwise comparison becomes a candidate strategy. 

\begin{table*}
\centering
\caption{Strategies found for Pacman and Dungeon Crawler, and the frequency with which we find each strategy, across 50 runs of our approach, excluding strategies with a found frequency below 60\%. Across all runs, both $\tau_P$ and $\tau_N$ contain 100 trajectories. A `*' indicates a grouping of strategies that have the same events, but with variations on the order of all but the final event. }
\label{table:1}
\begin{tabular}{l|l|l|l}
\toprule
    \textbf{Game $\mathcal{G}$}& \textbf{Event of Interest $E$} & \textbf{Strategies} & \textbf{Found(\%)} \\
    \midrule
    \multirow{4}{*}{Pacman} & kill a ghost & \textit{\{power-up, kill a ghost\}} &  100\\
    & kill a ghost $\times2$ & \textit{\{power-up, kill a ghost, kill a ghost\}} & 100 \\ 
    & kill a ghost $\times3$ & \textit{\{power-up, kill a ghost, kill a ghost, kill a ghost\}} & 100 \\ 
    \hline
    \multirow{4}{*}{Dungeon Crawler} & collect key & \textit{\{collect gun, collect key\}} & 100 \\
    & & \textit{\{collect sword, collect key\}} & 100 \\
    & & \textit{\{kill a monster, collect key\}} & 100 \\
    \cline{2-4}
    & kill a monster & \textit{\{collect gun, kill a monster\}} & 98 \\
    & & \textit{\{collect sword, kill a monster\}} & 98 \\
    & & \textit{\{collect key, kill a monster\}} & 88 \\
    \cline{2-4}
    & unlock door & \textit{\{collect sword, collect key, kill a monster, unlock door\}} & 100$^*$ \\
    & &\textit{\{collect gun, collect key, kill a monster, unlock door\}} & 94$^*$ \\
    & &\textit{\{collect key, unlock door\}} & 68 \\ 
    & &\textit{\{kill a monster, collect key, unlock door\}} & 66$^*$ \\
    & &\textit{\{collect gun, collect key, unlock door\}} & 60$^*$ \\
    \hline
    \multirow{2}{*}{Bank Heist} & destroy police car & \textit{\{rob bank, drop dynamite, destroy police car\}} & 100 \\
    & & \textit{\{rob bank, destroy police car\}} & 98 \\
    \cline{2-4}
    & rob bank $\times2$ & \textit{\{rob bank, drop dynamite, rob bank\}} & 100 \\
    & & \textit{\{rob bank, drop dynamite, destroy police car, rob bank\}} & 62 \\
    \cline{2-4}
    & rob bank $\times2$ and destroy police car & \textit{\{rob bank, drop dynamite, destroy police car, rob bank\}} & 100 \\
    \bottomrule
\end{tabular}
\end{table*}

\section{Experiments}

We evaluate the performance of our proposed method in the Pacman, Dungeon Crawler and Bank Heist games. For trajectory collection and determining events of interest, a PPO agent was trained using Masked PPO from \textit{stable\_baselines3}. Policies were trained for a maximum of 50,000 episodes. All experiments were performed on 3.2GHz Apple M1 with 16 GB RAM running Mac OS X 12.6.

To test the robustness of the extraction approach, we executed 50 runs for each environment under the same conditions and saved the resulting strategies and total counts of how many times each strategy was found. The results for all three games are shown in Table \ref{table:1}, focussing on strategies that were detected in at least 60\% of runs. In the Pacman environment, the predominant strategies, those with the highest `Found' percentage, are what we expected for each event of interest. Similarly, in the Bank Heist environment, the predominant strategies are reasonable for the corresponding events. 

In the Dungeon Crawler environment, we observed a combination of expected and unexpected strategies. In particular, many possible strategies were obtained for the ``unlock door'' event of interest. In this environment, the agent performs best by both killing the monster and unlocking the door. Consequently, many positive trajectories for both events are likely to involve both of these outcomes. This is evidenced by the event likelihoods computed for this event of interest, as shown in Table \ref{table:2}. Although the ``collect key'' event has the highest likelihood for the ``unlock door'' event of interest, the ``kill a monster'' event appears, on average, 75\% more in the positive trajectories than the negative. 

\begin{table}
\centering
\caption{Average event likelihoods for $E=$ \textit{``unlock door"} in the \textit{Dungeon Crawler} environment. Averages are calculated over all 50 experiment runs.}
\label{table:2}
\begin{tabular}{l|l}
\toprule
    \textbf{Event} & \textbf{Likelihood}\\
    \textbf{} & \textbf{(average(min,max))}\\
    \midrule
    ``collect key" & 0.94(0.89,0.96),\\
    ``kill a monster" & 0.75(0.67,0.79),\\
    ``collect sword" & 0.66(0.58,0.74),\\
    ``collect gun" & 0.65(0.6,0.73)\\
    \bottomrule
\end{tabular}
\end{table}

We also investigated the effect of the trajectory sample size i.e., how many trajectories are in the positive and negative sets, on the strategies extracted. Table \ref{table:3} shows that the expected strategies are still obtainable in the Dungeon Crawler environment after reducing the sample size. We observe that the sample size has an influence on the likelihood values computed for each event. We expect that likelihood values for certain events will become more accurate with increased sample size. The results of this analysis are shown in Table \ref{table:4}. 

\begin{table}
\centering
\caption{Strategies found across 50 runs of the extraction approach for the \textit{Dungeon Crawler} environment, using a sample size of 20 trajectories in $\tau_P$ and $\tau_N$.}
\label{table:3}
\begin{tabular}{l|l|l}
\toprule
    \textbf{Event of} & \textbf{Predominant} & \textbf{Found} \\
    \textbf{Interest} $E$ & \textbf{Strategies} &  \textbf{(\%)}\\
    \midrule
    collect & \textit{\{collect gun, collect key\}} &  88\\     
    key & \textit{\{collect sword, collect key\}} &  86\\
    \hline
    kill & \textit{\{collect sword, kill a monster\}} & 98\\
    a monster & \textit{\{collect gun, kill a monster\}} &  94\\
    \hline
    unlock  & \textit{\{collect key, unlock door\}} &  34\\ 
    door& \textit{\{collect sword, unlock door\}} &  24\\ 
    \bottomrule
\end{tabular}
\end{table}

\begin{table}
\centering
\caption{Average event likelihoods for $E=$ \textit{``kill a monster"} in the \textit{Dungeon Crawler} environment. Averages are calculated over 50 runs for each sample size.}
\label{table:4}
\begin{tabular}{l|l}
\toprule
    \textbf{Sample} & \textbf{Event Likelihoods}\\
    \textbf{Size} & \textbf{(average(min,max))}\\
    \midrule
    \multirow{3}{*}{10} & ``collect sword": 0.96(0.6,1),\\
    & ``collect gun": 0.91(0.6,1),\\
    & ``collect key": 0.75(0.1,1)\\
    \hline
    \multirow{3}{*}{50} & ``collect sword": 0.87(0.64,1),\\ 
    & ``collect gun": 0.77(0.6,1),\\
    & ``collect key": 0.19(0.12,0.36)\\
    \hline
    \multirow{3}{*}{100} & ``collect sword": 0.70(0.59,0.81),\\ 
    & ``collect gun": 0.69(0.61,0.78),\\
    & ``collect key": 0.22(0.13,0.34)\\
    \hline
    \multirow{3}{*}{200} & ``collect sword": 0.70(0.63,0.76),\\
    & ``collect gun": 0.69(0.6,0.78),\\
    & ``collect key": 0.22(0.17,0.3)\\
    \bottomrule
\end{tabular}
\end{table}

During this investigation, we observed a positive correlation between the average time taken to run all the stages for each event of interest, $\mathcal{T}$, and the sample size of the trajectories. This is due to the trajectory collection stage. For example, for the ``kill a monster'' event, stage (i) took, on average, 4 seconds for a sample size of 10 trajectories versus 100 seconds when the sample size is 200 trajectories. The time taken to complete stages (ii) to (v) was generally observed to be less than 1\% of $\mathcal{T}$, regardless of the sample size. 
 
Our results demonstrate that the proposed extraction approach is able to identify reasonable strategies for multiple events of interest. The performance of our approach is dependent on the ability to collect the same amount of positive and negative trajectories. In particular, in Table \ref{table:2}, for $E=$ \textit{``unlock door"} the strategy \textit{\{collect key, unlock door\}} is only found in 34\% of runs. This result is due to the limitations of the trajectory collection stage where events of interest were eventually discarded if not enough positive trajectories could be collected within a specified time limit. The second key result is that the event likelihood values are influenced by the sample size of trajectories. This is most noticeable in Table \ref{table:3}, the likelihood of ``collect key" to be part of a \textit{``kill a monster"} strategy decreases dramatically as sample size increases from 20 to 50. 

\section{Conclusion and Future Work}

We have proposed an approach for the extraction of strategies from learned agent policies as a first step toward improving the ability of artificial agents to transfer and generalise their existing knowledge. We adapted the Smith-Waterman local alignment algorithm to find useful causal information from agent trajectories which we can use to form strategies. Our results when demonstrated on video game trajectories obtained using reinforcement learning, showcase the ability of this method to identify reasonable strategy candidates in different contexts. 

In future work, we will utilise the strategies obtained from this method and look at generalisation techniques to support transfer to domains with differing action and state spaces and environmental dynamics. In particular, we will consider generalisation via abstraction, changing only the content of a strategy when lifting it to a more general context, leaving us with flexibility in the choice of data structures used. One approach is to use ontologies to determine how environment-specific events within the strategies can be lifted to a high-level representation that applies to the current environment. Abstracted strategies can be applied by leveraging functional transfer learning methods, including reward shaping \cite{gnm:19} and policy reuse \cite{bhtn:15}. Further, the proposed strategy extraction method is not limited to reinforcement learning agents. Therefore, a future research direction could be to consider strategy transfer in agents with different learning mechanisms. 

Ultimately, we envision this work could address issues around generalisability present in current state-of-the-art autonomous artificial agents and make them deployable in real-world scenarios. 


\bibliographystyle{ACM-Reference-Format} 
\bibliography{submission}

\balance

\end{document}